\def\year{2021}\relax
\documentclass[letterpaper]{article} % DO NOT CHANGE THIS
\usepackage{aaai21}  % DO NOT CHANGE THIS
\usepackage{times}  % DO NOT CHANGE THIS
\usepackage{helvet} % DO NOT CHANGE THIS
\usepackage{courier}  % DO NOT CHANGE THIS
\usepackage[hyphens]{url}  % DO NOT CHANGE THIS
\usepackage{graphicx} % DO NOT CHANGE THIS
\usepackage{natbib}  % DO NOT CHANGE THIS AND DO NOT ADD ANY OPTIONS TO IT
\usepackage{caption} % DO NOT CHANGE THIS AND DO NOT ADD ANY OPTIONS TO IT
\usepackage{amssymb}% http://ctan.org/pkg/amssymb
\usepackage{kotex} % korean
\usepackage{tabularx} % in the preamble
\usepackage{subfigure}
\usepackage{verbatim}  % comment
\usepackage{adjustbox}
\usepackage{url}
\usepackage{amsmath, soul}
\usepackage[ruled]{algorithm2e}
\usepackage{comment}
\usepackage[pagebackref=true,breaklinks=true,letterpaper=true, bookmarks=false]{hyperref}

\usepackage{color, colortbl}
\definecolor{Gray}{gray}{0.9}
\definecolor{LightCyan}{rgb}{0.88,1,1}

\title{Discriminative Region Suppression for \\Weakly-Supervised Semantic Segmentation}
\author{

    Beomyoung Kim, Sangeun Han, Junmo Kim
    \\
}
\affiliations{
    Korea Advanced Institute of Science and Technology (KAIST) \\
    \{qjadud1994, bichoomi, junmo.kim\}@kaist.ac.kr
}

\begin{document}

\maketitle

\begin{abstract}
Weakly-supervised semantic segmentation (WSSS) using image-level labels has recently attracted much attention for reducing annotation costs.
Existing WSSS methods utilize localization maps from the classification network to generate pseudo segmentation labels.
However, since localization maps obtained from the classifier focus only on sparse discriminative object regions, it is difficult to generate high-quality segmentation labels.
To address this issue, we introduce discriminative region suppression (DRS) module that is a simple yet effective method to expand object activation regions.
DRS suppresses the attention on discriminative regions and spreads it to adjacent non-discriminative regions, generating dense localization maps.
DRS requires few or no additional parameters and can be plugged into any network.
Furthermore, we introduce an additional learning strategy to give a self-enhancement of localization maps, named localization map refinement learning.
Benefiting from this refinement learning, localization maps are refined and enhanced by recovering some missing parts or removing noise itself.
Due to its simplicity and effectiveness, our approach achieves mIoU 71.4\% on the PASCAL VOC 2012 segmentation benchmark using only image-level labels.
Extensive experiments demonstrate the effectiveness of our approach.
The code is available at \url{https://github.com/qjadud1994/DRS}.
\end{abstract}

%------------------------------------------------------------------------------------
\section{Introduction}
\label{sec:intro}
Recent developments in deep learning have achieved great success on semantic segmentation tasks with the help of deep convolutional neural networks (CNNs) and rich pixel-level annotations.
However, collecting a large-scale pixel-level annotated dataset requires intensive human labor, which is both expensive and time-consuming.
To end this limitation, weakly-supervised semantic segmentation (WSSS) using only image-level labels has recently attracted much attention.

\begin{figure}[t]
    \centering

    \includegraphics[width=\linewidth]{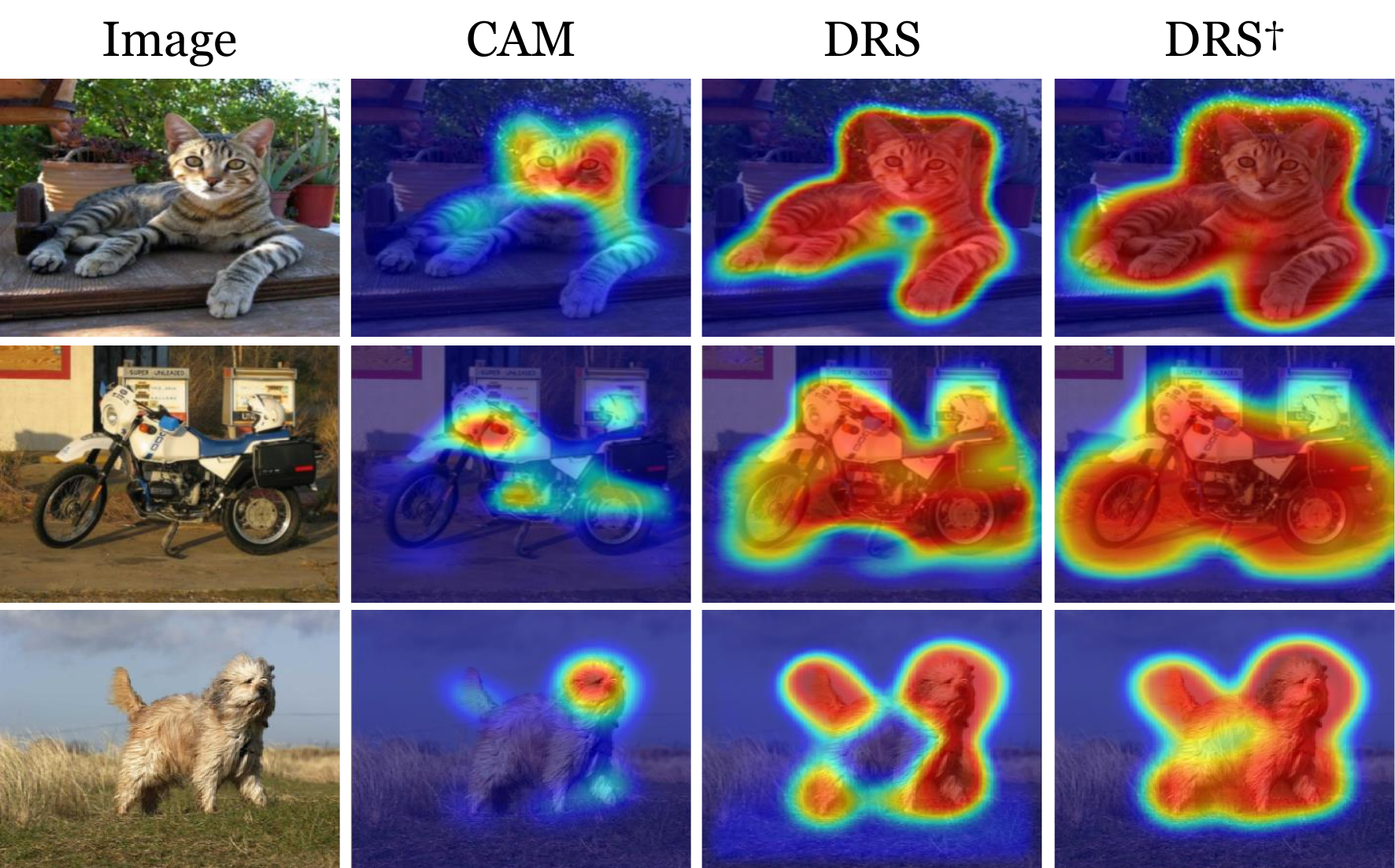}
    \caption{Visual comparisons of localization maps produced by original CAM, DRS, and DRS$\dagger$. DRS$\dagger$ denotes that the refinement learning is applied.}
    \label{figure:concept}
\end{figure}

One problem with using image-level annotations is that we have no information about the location of the target object; we only know whether the object is present in the image or not.
This makes semantic segmentation learning challenging.
To learn pixel-level semantic knowledge from image-level labels, it is common practice to use localization maps obtained from the classification network using class activation maps (CAMs)~\cite{zhou2016learning}. 
Specifically, the discriminative region for each target class provided by CAMs is used as pixel-level supervision for segmentation network training.
However, this discriminative region is usually very sparse and only covers a small part of the object, which is not enough for semantic segmentation learning as shown in the second column in Figure \ref{figure:concept}. 
Therefore, most studies in the weakly-supervised semantic segmentation field focus on expanding the object region to produce dense localization maps.
One of the recent approaches is image-level and feature-level erasure of discriminative parts~\cite{wei2017object, li2018tell, hou2018self}.
This approach strictly erases discriminative parts, letting the network focus on other non-discriminative parts.
However, they not only tend to produce undesired true negative regions when most of the discriminative parts are erased but also require a lot of additional parameters for multiple classifiers or multiple branches.

%------------------------------------------------------------------------------------
%------------------------------------------------------------------------------------

%------------------------------------------------------------------------------------
\begin{figure*}[t]
    \centering
    \includegraphics[width=\linewidth]{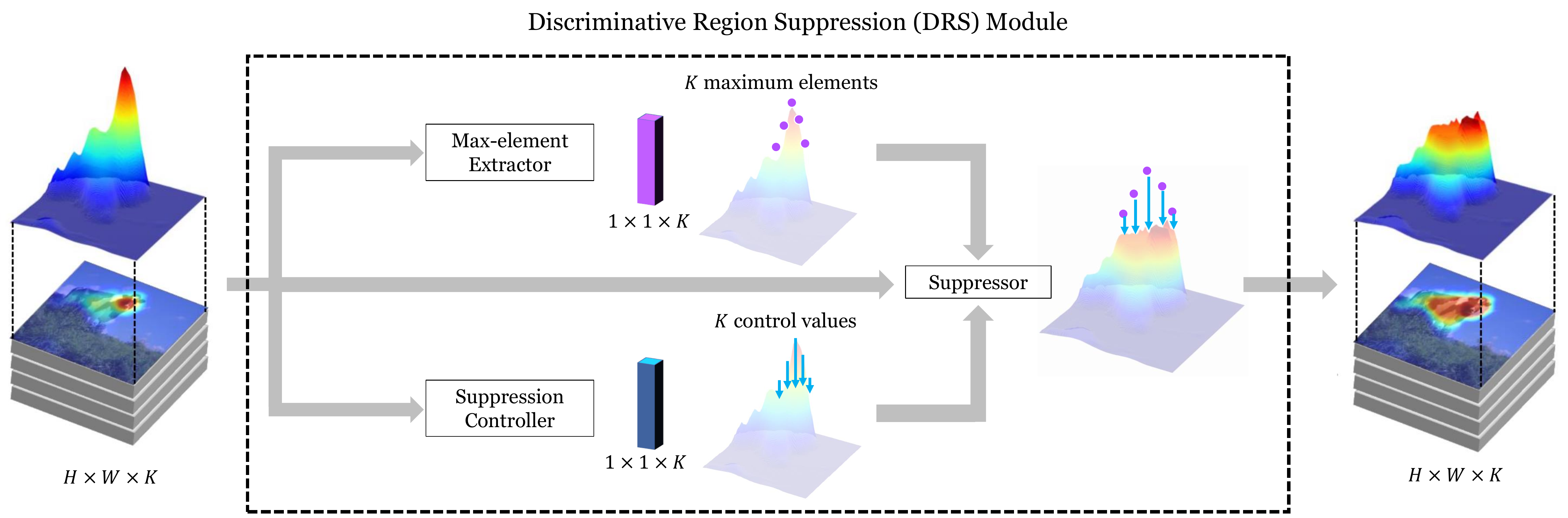}
    \caption{
        Diagram of our discriminative region suppression (DRS) module. DRS suppresses intermediate feature maps, diffusing the attention into adjacent non-discriminative parts.
        The max-element extractor extracts $K$ maximum elements from intermediate feature maps. 
        These $K$ maximum elements are the maximum points of each discriminative region and are considered as starting points to be suppressed.
        For convenience, $K$ maximum elements are illustrated in 5 purple points.
        The controller predicts control values, which determine how much to suppress feature maps from these $K$ maximum elements.
        These $K$ control values are illustrated in 5 blue arrows and the length of the arrow means how much suppress feature maps from the corresponding maximum element.
        Using these $K$ maximum elements and $K$ control values, the suppressor suppresses discriminative regions and spreads the attention into adjacent non-discriminative parts.
    }
    \label{figure:main}
\end{figure*}
%------------------------------------------------------------------------------------

In this paper, we propose discriminative region suppression (DRS) module, which is a simple and efficient yet effective and novel approach for generating dense localization maps.
The goal of DRS is to suppress discriminative regions, not to erase them, so that attention spreads to adjacent non-discriminative regions; this mild approach helps the classifier effectively expand discriminative object regions.
DRS module consists of three components as depicted in Figure \ref{figure:main}: max-element extractor, suppression controller, and suppressor.
These components work together to produce dense localization maps by reducing the attention gap between discriminative regions and adjacent non-discriminative regions.
DRS not only effectively expands the object regions without generating much noise, but also can be plugged into any network with few or no additional parameters.

Although we can obtain dense segmentation labels from the classification network equipped with DRS, it does not recover missing parts or weak attention by itself because the objective of the classification network is classification, not localization.
To address this issue, we introduce an additional training strategy, named localization map refinement learning, inspired by \cite{jiang2019integral}
Localization map refinement learning induces self-enhancement of localization maps by recovering missing or weak attention region.
In Figure \ref{figure:concept}, we compare some results of DRS and DRS$\dagger$, where DRS$\dagger$ denotes that refinement learning is applied.

Following the convention, we generate pseudo segmentation labels from our dense localization maps and evaluate on weakly-supervised semantic segmentation task.
On the PASCAL VOC 2012 segmentation benchmark, we achieve mIoU 71.4\% on the testset using only image-level labels.
In addition, extensive experiments demonstrate the effectiveness of our approach.

In summary, the contributions of our work are as follows:
\begin{itemize}
\item We introduce a simple, effective, and novel approach for weakly-supervised semantic segmentation named discriminative region suppression (DRS) module, which requires few or no additional parameters and can be easily plugged into any network.
\item DRS effectively and efficiently suppresses discriminative regions to generate dense localization maps, bridging the gap between discriminative regions and adjacent non-discriminative regions. 
\item For self-enhancement of localization maps, we introduce an additional training strategy, named localization map refinement learning.
\item Extensive experiments and analyses demonstrate the effectiveness of our DRS module and we achieve competitive performance on Pascal VOC 2012 segmentation benchmark using only image-level labels.
\end{itemize}

\section{Related Work}

Most recent studies on semantic segmentation using image-level labels as weak supervision utilize CAMs \cite{zhou2016learning} to localize object regions and focus on expanding them to non-discriminative parts of the objects.
To this end, AE-PSL \cite{wei2017object}, GAIN \cite{li2018tell}, and SeeNet \cite{hou2018self} propose erasing techniques to generate dense localization maps. However, these erasure-based approaches usually require multiple classifiers and complicated training procedures. 
Moreover, erasing most of the discriminative regions may introduce true negative regions and confuse the classifier.

To avoid the repetitive training procedures of AE-PSL \cite{wei2017object}, MDC \cite{wei2018revisiting} propose a multi-dilated convolution block in which the receptive fields of various sizes capture different patterns.
As a more generalized approach, FickleNet \cite{lee2019ficklenet} aggregate diverse localization maps produced by stochastic feature selection. 
Although they effectively expand the activated regions, some falsely labeled regions outside the object tend to be identified because the receptive fields of these methods are not adaptive to object size.
The recently proposed OAA \cite{jiang2019integral} accumulates attention maps at different training epochs and introduces integral attention learning to enhance attention maps.
However, it may produce undesired attention regions due to training instability in the early stage. 

Some other works \cite{ahn2018learning, huang2018weakly, shimoda2019self} adopt a region-growing technique to expand initial regions. 
More recently, RRM~\cite{zhang2020reliability} proposed a fully end-to-end network for joint training of classification and segmentation, and SGAN~\cite{yao2020saliency} proposed a self-attention network guided by saliency priors that can produce dense and accurate localization maps from rich contextual information.
BES~\cite{chen2020weakly} explores object boundaries to refine the semantic segmentation output.
ICD~\cite{fan2020learning} proposes an intra-class discriminator approach to separate foreground objects and the background within the same image-level class.

%------------------------------------------------------------------------------------
\begin{figure*}[t]
    \centering
    \includegraphics[width=0.78\linewidth]{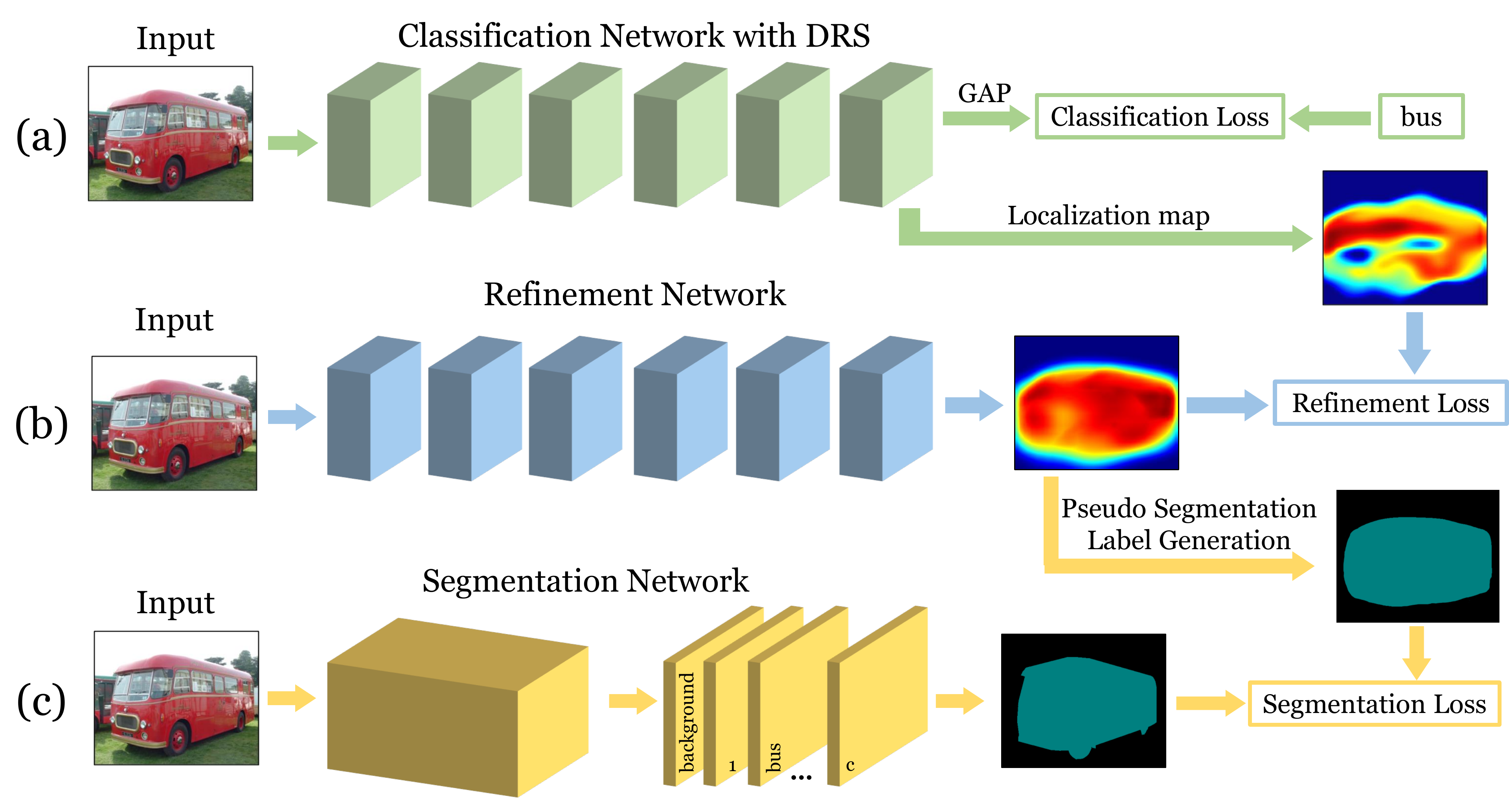}
    \caption{
        Overview of the proposed method. (a) Classification network with DRS for obtaining localization maps, (b) localization map refinement learning, and (c) weakly-supervised semantic segmentation.
        (a), (b), and (c) are executed sequentially, not simultaneously. Note that GAP in (a) means the global average pooling layer.
    }
    \label{figure:network}
\end{figure*}
%------------------------------------------------------------------------------------

\section{Method}

The overview of our method is illustrated in Figure \ref{figure:network}.
We sequentially train three different networks for classification, refinement, and segmentation.
After training the classification network with the discriminative region suppression (DRS) module, we produce dense localization maps.
Using these dense localization maps as ground truth labels for refinement learning, we train the refinement network to produce refined localization maps.
Then, pseudo segmentation labels are generated from the refined localization maps and used for training the semantic segmentation network.
We measure the segmentation performance to evaluate the quality of our localization maps.

\subsection{Observation}
\label{method:1}
We produce localization maps from the class-specific feature maps of the last convolutional layer, which have been proven by \cite{zhang2018adversarial} to be mathematically equivalent to CAMs.
We employ the VGG-16~\cite{simonyan2014very} as our classification network to produce localization maps.
To be specific, we employ modified VGG-16, where all fully connected layers are removed. On top of it, three convolutional layers with 512 channels and kernel size 3, and a convolutional layer with $C$ channels and kernel size 1 are added. Here $C$ is the number of categories.
This network produces output feature maps $F \in \mathbb{R}^{H_{out} \times W_{out} \times C}$ and classification score $P=\sigma(GAP(F))$ from input image. 
$H_{out}$ and $W_{out}$ are the height and width of output feature maps, respectively; $GAP(\cdot)$ is the global average pooling; and $\sigma(\cdot)$ is the sigmoid function.
For each target category $c$, $c$-th localization map $M^{c}$ is defined as the normalized $c$-th feature map $F^{c}$:
\begin{equation}
    M^c=\frac{ReLU(F^c)}{max(F^c)}.
\end{equation}

From the definition of $M$, we observe that discriminative object regions are identified with relatively high values on the feature maps $F$.
Based on this observation, we regard the high-value areas on feature maps as discriminative regions.

%---------------------------------------------------------------------
\begin{algorithm}[t]
    \SetKwInOut{Input}{Input}
        \Input{
            Intermediate feature maps $X \in R^{H \times W \times K}$
        }
    {
    \SetKwInOut{Output}{Output}
        \Output{
            Suppressed feature maps : $\hat{X} \in R^{H \times W \times K}$
        }
    \vspace{3mm}
    $X_{max}$ $\gets$ $extractor(X)$ \hspace{3mm}//$X_{max} \in \mathbb{R}^{1 \times 1 \times K}$ \\
    \vspace{2mm}
    $G$ $\gets$ $controller(X)$ \hspace{8mm}//  $G \in [0, 1]^{1 \times 1 \times K}$ \\
    \vspace{1mm}
    $\tau$ $\gets$ $X_{max} \cdot G$ \hspace{12mm}// upper bound, $\tau \in \mathbb{R}^{1 \times 1 \times K}$ \\
    \vspace{2mm}
    $\tau$ $\gets$ \textit{expand to the same shape of $X$} \\
    \vspace{2mm}
    $\hat{X}$ $\gets$ $min(X, \tau)$ \hspace{13mm}// suppressor \\
    }
\caption{Discriminative Region Suppression}
\label{DRS_algorithm}
\end{algorithm}
%---------------------------------------------------------------------

\subsection{Discriminative Region Suppression}
\label{method:2}
To produce dense localization maps, we propose discriminative region suppression (DRS) module.
The main problem of segmentation label generation using CAMs is that discriminative regions only appear partially and sparsely, as shown in the second column of Figure~\ref{figure:concept}.
To address this issue, DRS aims to spread the attention on discriminative regions to adjacent non-discriminative regions.
Specifically, DRS suppresses the attention on discriminative regions, allowing the network to focus on non-discriminative regions.

Let $X \in \mathbb{R}^{H \times W \times K}$ be an intermediate feature map, where $H$, $W$, and $K$ are the height, width, and the number of channels of $X$.
DRS module consists of three parts: max-element extractor, suppression controller, and suppressor.
The max-element extractor extracts $K$ maximum elements from the intermediate feature map $X$ using global max pooling.
The output of the extractor is denoted as $X_{max} \in \mathbb{R}^{1 \times 1 \times K}$. 
Based on the observation, these $K$ maximum elements are regarded as the criteria of discriminative regions and considered as starting points to be suppressed.

The suppression controller determines how much to suppress discriminative regions.
In detail, it generates $G \in [0, 1]^{1 \times 1 \times K}$ and each $k$-th control value in $G$ determines the amount of suppression in $X$ with respect to the corresponding $k$-th maximum element.

Using the $K$ maximum elements and $K$ control values, the suppressor suppresses discriminative regions. Specifically, element-wise multiplication of $X_{max}$ and $G$ is regarded as the upper bound of $X$, denoted as $\tau = X_{max} \cdot G$, $\tau \in \mathbb{R}^{1 \times 1 \times K}$.
The regions in $X$ above this upper bound are regarded as discriminative regions to be suppressed. After the upper bound $\tau$ is expanded to the same shape of $X$, the element-wise minimum operation is applied on $X$ and $\tau$ to suppress discriminative regions. 
For example, if the $k$-th control value is 0.7, $X^k$ is suppressed until no element exceeds 70\% of the $k$-th maximum value.
In this way, the suppressor bridges the gap between discriminative regions and adjacent non-discriminative regions.
The whole process of DRS is described in Algorithm~\ref{DRS_algorithm} and illustrated in Figure~\ref{figure:main}.

For the suppression controller, there are two types of controller: learnable controller and non-learnable controller.
If the suppression power is too strong, the discriminative feature extraction power is weakened.
The learnable controller adaptively balances between discriminative feature extraction power and suppression power of the classification network. 
Formally, the output of the learnable controller is
\begin{equation}
    G = \sigma(f(GAP(X);\theta)), 
\end{equation}
where $f$ is a fully connected layer, $\theta$ is a learnable parameter of the controller, and $G \in [0, 1]^{1 \times 1 \times K}$.
Since $\theta$ is trained with the classification objective, DRS with a learnable controller adaptively suppresses 
discriminative regions so as not to damage the discriminative feature extraction power much.

To produce even more dense localization maps at the expense of discriminative feature extraction power, we forcibly suppress discriminative regions; this is the goal of a non-learnable controller.
For the non-learnable controller, each element of $G$ is set to a constant value $\delta$.
We set the hyperparameter $\delta$ to a value between 0 and 1, and a lower $\delta$ means more intense suppression resulting in more dense localization maps.
Compared to the learnable controller, the non-learnable controller does not require additional training parameters but requires a hyperparameter $\delta$.
In the experiment section, we analyze both learnable and non-learnable controller with quantitative and qualitative results.

Figure \ref{figure:network} (a) illustrates the process of obtaining dense localization maps from the classification network with DRS.
As shown in the third column of Figure~\ref{figure:concept}, DRS reduces the gap between the activation of discriminative regions and adjacent non-discriminative regions to obtain dense localization maps. 
Note that DRS can be plugged into any layer of a network.

\subsection{Localization Map Refinement Learning} 
\label{method:3}
Although DRS helps produce dense localization maps, the DRS itself lacks the ability to recover missing parts of the target objects or enhance weak attention in adjacent non-discriminative regions because the goal of the classification network is essentially classification ability, not localization map generation.
Motivated by \cite{jiang2019integral}, we introduce an additional learning strategy for localization map refinement to solve the above limitations.
This learning strategy for self-enhancement of localization maps is called localization map refinement learning, denoted as DRS$\dagger$.
After training the classification network with DRS, we exploit the output localization maps $M \in [0,1]^{H_{out} \times W_{out} \times C}$ as the ground truth localization maps for refinement learning.	

The network for refinement learning, called refinement network, is based on the VGG-16; all fully-connected layers are removed and three convolutional layers with 512 channels and kernel size 3, and a convolutional layer with C channels and kernel size 1 are appended.
The refinement network directly produces refined localization maps $N \in \mathbb{R}^{H_{out} \times W_{out} \times C}$, which have the same shape as $M$.
We adopt the mean squared error (MSE) loss function as the refinement loss for the refinement network.
Refinement learning is depicted in Figure \ref{figure:network} (b).

Benefiting from refinement learning, we can obtain more dense and high-quality localization maps through self-enhancement, as shown in Figure~\ref{figure:concept} (DRS$\dagger$ $v.s.$ DRS)

%------------------------------------------------------------------------------------
\begin{figure*}[t]
    \centering
    \includegraphics[width=\linewidth]{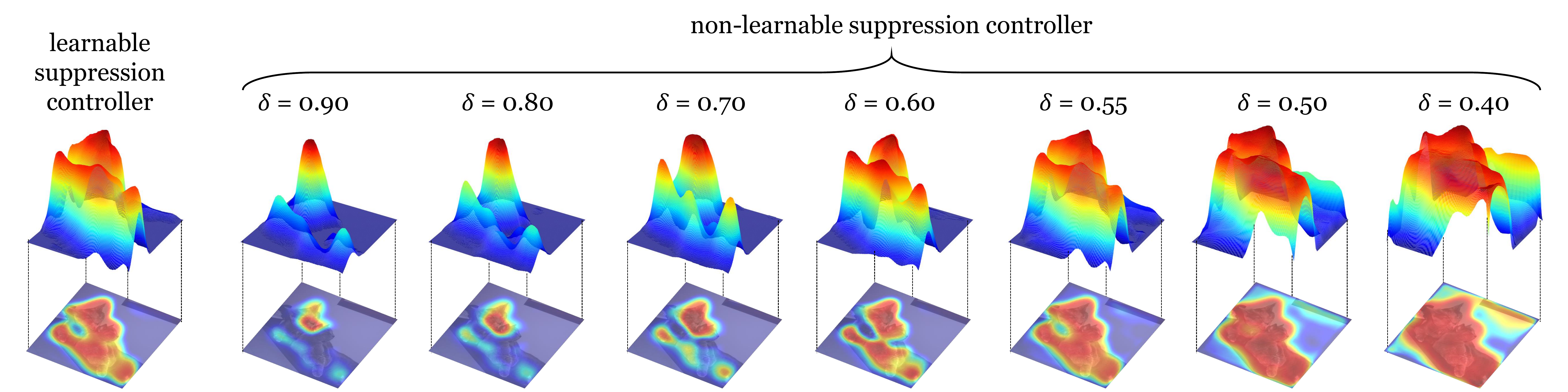}
    \caption{
        Visualization of localization maps of two types of controllers and the non-learnable controller with different $\delta$.
    }
    \label{figure:controller}
\end{figure*}
%------------------------------------------------------------------------------------

\subsection{Weakly-Supervised Semantic Segmentation}
\label{method:4}
Using our dense localization maps obtained from the refinement network, we generate pseudo segmentation labels and use them as weak-supervision for the semantic segmentation network.
We generate pseudo segmentation labels using object cues and background cues.
We extract object cues from the localization maps by taking the pixels whose values are higher than $\alpha$ and extract background cues using salient object detection method~\cite{liu2019simple}, motivated by \cite{wei2017object, wei2018revisiting}; the pixels with saliency values lower than $\beta$ are taken as background.
Those who belong to neither of the cues are ignored.
  
Following the convention, we train the segmentation network such as \cite{chen2014semantic, chen2017deeplab} using the generated pseudo segmentation labels, as illustrated in Figure \ref{figure:network} (c).
The segmentation performance is compared with other methods using the same segmentation network to evaluate the quality of pseudo segmentation labels.

\section{Experiments}

\subsection{Dataset and Evaluation Metrics}
    
We demonstrate the effectiveness of the proposed approach on the PASCAL VOC 2012 segmentation benchmark dataset \cite{Everingham2014ThePV}, which contains 20 object categories and one background category. 
Following the common practice in previous works, the training set is augmented to 10,582 images.
We evaluate the performance of our model using the mean intersection-over-union (mIoU) metric and compare it with other state-of-the-art methods on the validation (1,449 images) and test set (1,456 images).
For the test results, we submit the prediction outputs to the official PASCAL VOC evaluation server.

\subsection{Implementation details}

For the classification network, we adopt the modified VGG-16 with DRS plugged into every layer, as mentioned in the method section.
Its parameters are initialized by the VGG-16~\cite{simonyan2014very} pre-trained on ImageNet~\cite{deng2009imagenet} except for the additional convolutional layers.
We train the classification network with binary cross-entropy loss using the SGD optimizer with a weight decay of 5e-4 and a momentum of 0.9. 
The initial learning rate is set to 1e-3 and is decreased by a factor of 10 at epoch 5 and 10. 
For data augmentation, we apply a random crop with 321$\times$321 size, random horizontal flipping, and random color jittering.
We use a batch size of 5 and train the classification network for 15 epochs.

We optimize the refinement network for the refinement learning with MSE loss using Adam~\cite{kingma2014adam} optimizer with a learning rate of 1e-4. The batch size is 5, the total training epoch is 15, and the learning rate is dropped by a factor of 10 at epoch 5 and 10.
We apply the data same augmentation strategy as in the classification network.

For the segmentation network, we experiment with three architectures: DeepLab-Large-FOV~\cite{chen2014semantic} with VGG-16 and DeepLab-Large-FOV with ResNet-101~\cite{he2016deep} backbones, and DeepLab-ASPP~\cite{chen2017deeplab} with ResNet-101 backbone.
When generating pseudo segmentation labels, we empirically choose $\alpha = 0.2$ for object cues and $\beta = 0.06$ for background cues.
Our method is implemented on Pytorch \cite{paszke2017automatic}.
We use the DeepLab-Large-FOV code\footnote{\url{https://github.com/wangleihitcs/DeepLab-V1-PyTorch}} and DeepLab-ASPP code\footnote{\url{https://github.com/kazuto1011/deeplab-pytorch}} implemented based on the Pytorch framework, following the same hyperparameter settings for training and the conditional random field (CRF)~\cite{krahenbuhl2011efficient} as the original publications.
All experiments are performed on NVIDIA TITAN XP.

\begin{comment}
%------------------------------------------------------------------------------------
\begin{table}[t]
  \centering
  \caption{
    Effect of the two types of controllers and the hyperparameter $\delta$ for the non-learnable controller.
  }
  \label{tab:controller}
  \begin{adjustbox}{max width=\linewidth}
            \begin{tabular}{c|ccccccc}
            \hline
            \multicolumn{8}{c}{suppression controller} \\
            \hline
            learnable & \multicolumn{7}{c}{non-learnable} \\
                      & $\delta$=0.90 & $\delta$=0.80 & $\delta$=0.70 & $\delta$=0.60 & $\delta$=0.55 & $\delta$=0.50 & $\delta$=0.40 \\
            \hline
            \textbf{62.9\%} & 51.9\% & 56.0\% & 58.7\% & 62.3\% & \textbf{62.8\%} & 62.3\% & 59.6\% \\
            
            \hline
          \end{tabular}
        \end{adjustbox}
\end{table}
%------------------------------------------------------------------------------------
\end{comment}

%------------------------------------------------------------------------------------
\begin{table}[t]
  \centering
  \begin{adjustbox}{max width=\linewidth}
            \begin{tabular}{c|c}
            \hline
            suppression controller & mIoU \\
            \hline
            learnable & \textbf{62.9\%} \\
            non-learnable with $\delta$=0.90 & 51.9\% \\
            non-learnable with $\delta$=0.80 & 56.0\% \\
            non-learnable with $\delta$=0.70 & 58.7\% \\
            non-learnable with $\delta$=0.60 & 62.3\% \\
            non-learnable with $\delta$=0.55 & \textbf{62.8\%} \\
            non-learnable with $\delta$=0.50 & 62.3\% \\
            non-learnable with $\delta$=0.40 & 59.6\% \\
            \hline
          \end{tabular}
  \end{adjustbox}
  \caption{
    Effect of the two types of controllers and the hyperparameter $\delta$ for the non-learnable controller.
  }
  \label{tab:controller}
\end{table}
%------------------------------------------------------------------------------------

\subsection{Analysis}
To analyze the effectiveness of the proposed method, we conduct several experiments.
Following the convention of weakly-supervised semantic segmentation, we measure the mIoU score of the segmentation network outputs to evaluate the quality of our localization maps.
For all experiments in this section, we adopt the DeepLab-Large-FOV with VGG-16 as the segmentation network and measure the mIoU score on the VOC 2012 validation set.

%------------------------------------------------------------------------------------
\begin{figure}[t]
    \centering
    \includegraphics[width=\linewidth]{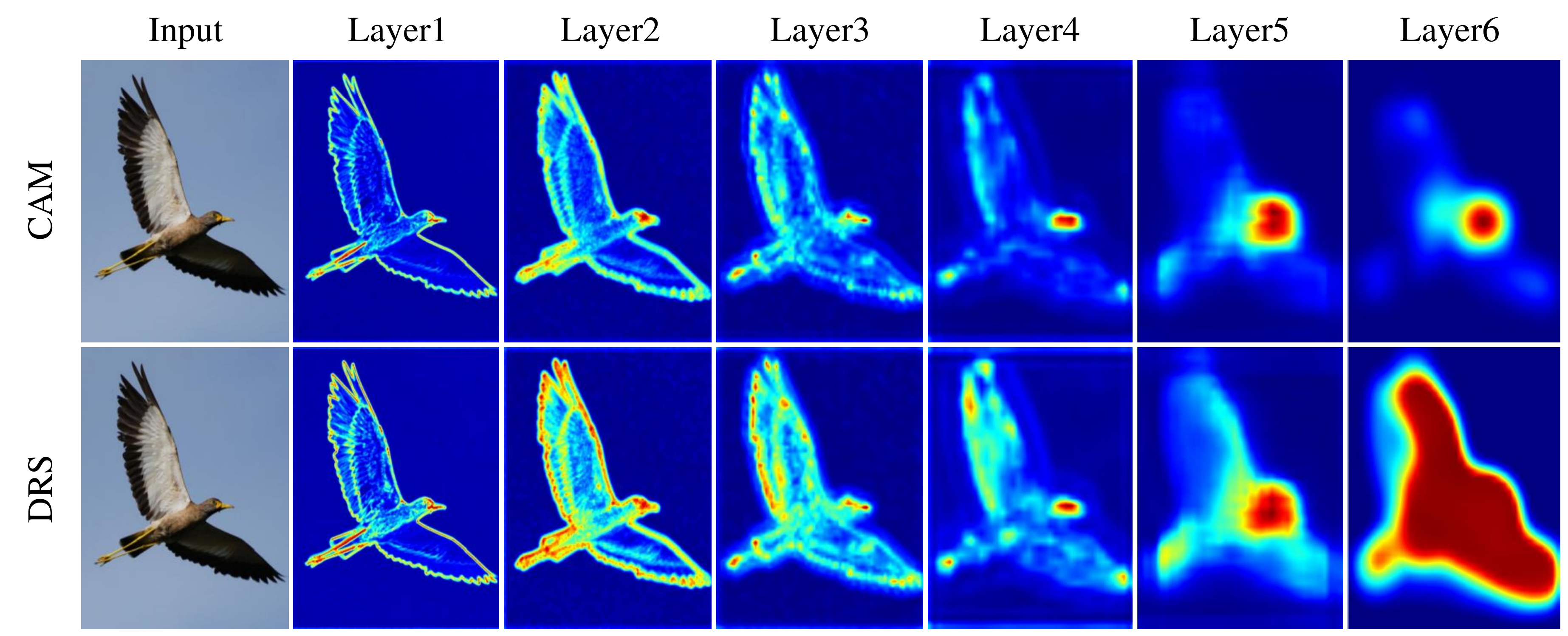}
    \caption{
        Visualization of feature maps on each layer. Note that element-wise averaging and normalization are applied to feature maps of each layer for visualization. 
    }
    \label{figure:activation}
\end{figure}
%------------------------------------------------------------------------------------

\noindent\textbf{Suppression controller.} 
In the method section, we introduced two types of suppression controller: learnable and non-learnable controller.
We investigate both controllers and the effect of the parameter $\delta$ for the non-learnable controller using visualization and quantitative analysis. For this analysis, we plugin the DRS to all layers of the classification network and skip the refinement learning procedure for precise effect analysis.
In the case of a non-learnable controller, we set the same $\delta$ for all layers.

Firstly, we analyze the effect of $\delta$ for the non-learnable controller and compare the output localization maps of each $\delta$ in Figure~\ref{figure:controller}.
When $\delta$ is 0.90, the localization map is mostly activated in the head of the cat.
Consequently, the gap between discriminative regions and adjacent non-discriminative regions is large, resulting in sparse localization maps.
As the $\delta$ gets smaller, activation at the body of the cat becomes higher, and the activation gap between the head and the body of the cat is smaller.
This indicates as the $\delta$ value decreases, the discriminative regions are further suppressed and the gap between discriminative regions and non-discriminative regions becomes smaller, resulting in dense localization maps.
However, if the $\delta$ value is too low ($i.e.,$ too much suppression), the gap between the background and the foreground becomes very small, resulting in a noisy localization map as shown in the rightmost result of Figure~\ref{figure:controller}.
Therefore, it is important to set an appropriate $\delta$ value for the non-learnable controller.
The quantitative results in Table \ref{tab:controller} support our arguments. 
The non-learnable controller with $\delta=0.55$ achieves better performance than that of $\delta=0.90$ (62.8\% $v.s.$ 51.9\%), but in the case of over-suppression, $e.g.$ the non-learnable controller with $\delta=0.40$, the performance is rather worse than that of $\delta=0.55$ (59.6\% $v.s.$ 62.8\%).
Through this experiment, we found that $\delta=0.55$ yields the best mIoU performance.

In the case of a learnable controller, it suppresses without generating much noise, creating moderately dense localization maps as in the leftmost of Figure~\ref{figure:controller}.
Compared to the non-learnable controller with $\delta=0.55$, the learnable controller produces similar mIoU performance (62.9\% $v.s.$ 62.8\%) and localization maps.
However, the classification accuracy of the learnable controller is much higher (72.6\% $v.s.$ 68.7\%).
From these results, we can notice that the learnable controller adaptively balances between the discriminative feature extraction power and the suppression power, whereas the non-learnable controller forcibly increases the suppression power at the expense of the feature extraction power.
Note that the learnable controller is free from hyperparameter ($i.e.$, $\delta$) tuning, but requires additional training parameters (from 21.8M to 24.4M training parameters).

%------------------------------------------------------------------------------------
\begin{table}[t]
  \centering
  \begin{adjustbox}{max width=\linewidth}
          \begin{tabular}{cccccc|c}
            \hline
            layer1 & layer2 & layer3 & layer4 & layer5 & layer6 & mIoU  \\
            
            \hline
            -           & -           & -           & -           & -           & -           &  50.1\%  \\
            -           & -           & -           & -           & -           & \checkmark  &  60.8\%  \\
            -           & -           & -           & -           & \checkmark  & \checkmark  &  62.2\%  \\
            -           & -           & -           & \checkmark  & \checkmark  & \checkmark  &  62.8\%  \\
            -           & -           & \checkmark  & \checkmark  & \checkmark  & \checkmark  &  \textbf{62.9\%}  \\
            -           & \checkmark  & \checkmark  & \checkmark  & \checkmark  & \checkmark  &  62.7\%  \\
            \checkmark  & \checkmark  & \checkmark  & \checkmark  & \checkmark  & \checkmark  &  \textbf{62.9\%}  \\
            \checkmark  & \checkmark  & \checkmark  & \checkmark  & \checkmark  & -           &  58.2\%  \\
            \checkmark  & \checkmark  & \checkmark  & \checkmark  & -           & -           &  53.6\%  \\
            \hline
          \end{tabular}
  \end{adjustbox}
  \caption{
    Effect of DRS in each layer. \checkmark means DRS is applied.
  }
  \label{tab:each_layer}
\end{table}
%------------------------------------------------------------------------------------

%------------------------------------------------------------------------------------
\begin{table}[t]
  \centering
  \begin{adjustbox}{max width=\linewidth}
            \begin{tabular}{c|c|c}
            \hline
            suppression controller & without refine & with refine  \\
            \hline
            learnable & 62.9\%  & 63.5\%\\
            non-learnable ($\delta$=0.55) & 62.8\% & 63.6\%  \\
            
            \hline
          \end{tabular}
  \end{adjustbox}
  \caption{
    Effect of localization map refinement learning.
  }
  \label{tab:refine}
\end{table}
%------------------------------------------------------------------------------------

\noindent\textbf{Effect of DRS on each layer.}
To observe the effect of DRS on each layer, we employ two analytical methods: visualization and quantitative analysis.
For this analysis, we use the DRS module with the learnable controller and skip refinement learning for precise effect analysis. 

For visualization, we apply element-wise averaging and normalization from 0 to 1 on feature maps of every layer. Figure \ref{figure:activation} shows the visualization results of the original CAM and our DRS-plugged classification network.
In lower-level layers ($i.e.$, from layer1 to layer3), we notice that the effect of DRS is minor because a network mainly focuses on the local features ($e.g.$, edge) where the gap between the discriminative and adjacent non-discriminative regions tends to be extremely large.
Meanwhile, in higher-level layers ($i.e.$, from layer4 to layer6), a network mostly focuses on the global features ($e.g.$, head of a bird) where the gap between the discriminative and adjacent non-discriminative regions is relatively small.
In this case, the effect of DRS becomes significant because it suppresses the activation of discriminative regions and expands the attention to non-discriminative regions.

For quantitative analysis, we plug in and out DRS at each layer and evaluate the performance of each case.
The results in Table \ref{tab:each_layer} show that the more we plugin the DRS at higher-level layers, the higher performance (from 50.1\% to 62.9\%).
On the other hand, applying DRS in lower-level layers has little effect (62.9\% $v.s.$ 62.7\%).
In addition, when we plug-out the DRS at higher-level layers, the performance significantly decreases (from 62.9\% to 58.2\% and 53.6\%).
From these results, we can conclude that DRS is more effective to produce dense localization maps when applied in higher-level layers.

\begin{figure}[t]
    \centering
    \includegraphics[width=\linewidth]{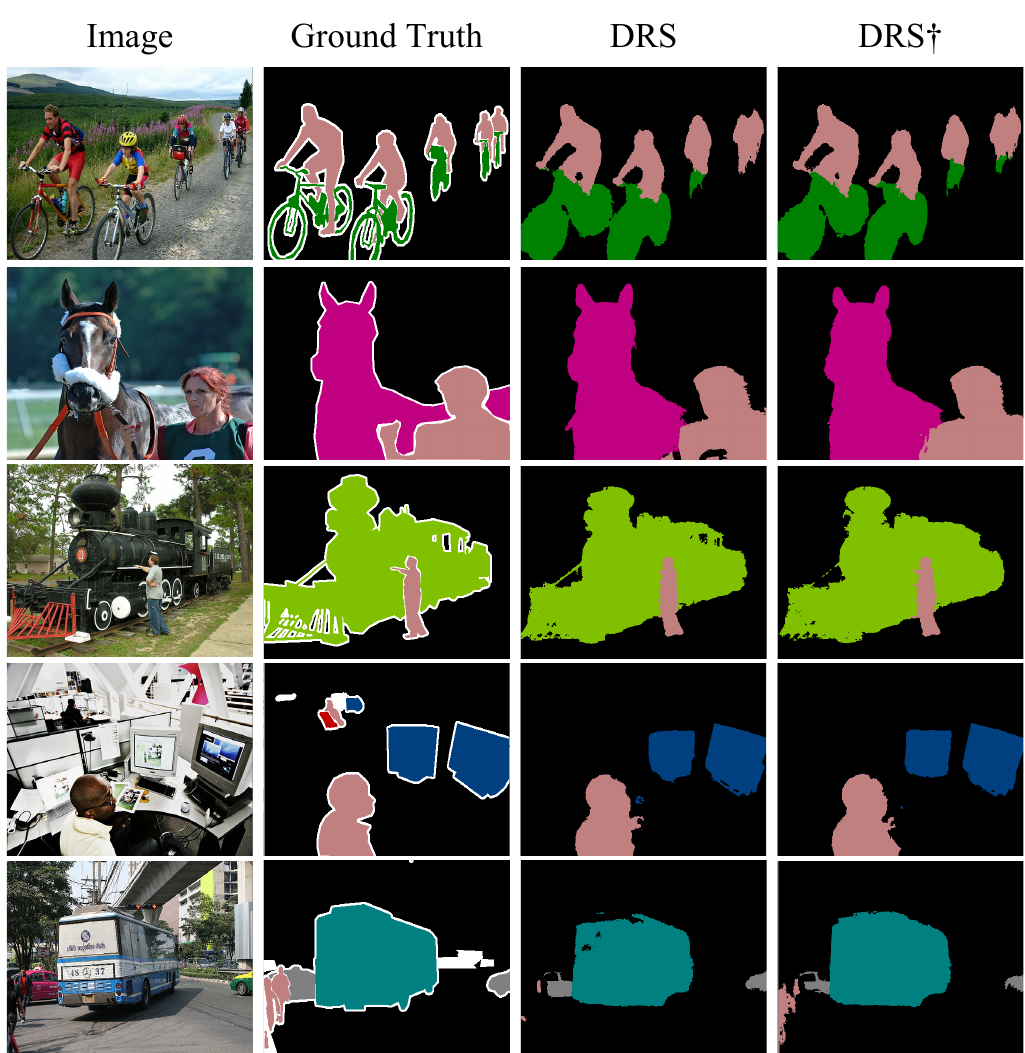}
    \caption{
        Qualitative results on the PASCAL VOC 2012 validation set.
    }
    \label{figure:result}
    \vspace{-1mm}
\end{figure}

\noindent\textbf{Improvement through refinement learning.}
As mentioned in the method section, localization map refinement learning gives a self-enhancement effect to produce high-quality dense localization maps as in Figure~\ref{figure:concept}. 
The improved mIoU performance is reported in Table~\ref{tab:refine} (+0.6\% and +0.8\%).
In addition, Figure \ref{figure:result} shows some segmentation results on the PASCAL VOC 2012, where both DRS and DRS$\dagger$ show satisfactory results, but DRS$\dagger$ leads to better segmentation results. 
Note that the learnable controller is used for Figure \ref{figure:concept} and \ref{figure:result}.

%------------------------------------------------------------------------------------
\begin{table}[t]
  \centering
  \begin{adjustbox}{max width=\linewidth}
  \begin{tabular}{ccccc}
    \hline
    Method & $\mathcal{S}$ & val & test \\
    \hline
    \multicolumn{4}{c}{\textbf{Segmentation Network : DeepLab-Large-FOV (VGG-16)}} \\
    \hline
    AE-PSL~\cite{wei2017object} & \checkmark & 55.0\% & 55.7\% \\
    GAIN~\cite{li2018tell}      & \checkmark & 55.3\% & 56.8\% \\ 
    MCOF~\cite{wang2018weakly}  & \checkmark & 56.2\% & 57.6\% \\
    %DCSP~\cite{chaudhry2017discovering}  & \checkmark & 58.6\% & 59.2\% \\
    AffinityNet~\cite{ahn2018learning} & -        & 58.4\% & 60.5\% \\
    SeeNet~\cite{hou2018self}       & \checkmark & 61.1\% & 60.7\% \\
    MDC~\cite{wei2018revisiting}    & \checkmark & 60.4\% & 60.8\% \\
    RRM~\cite{zhang2020reliability} &  -        & 60.7\% & 61.0\% \\
    FickleNet~\cite{lee2019ficklenet}  & \checkmark & 61.2\% & 61.8\% \\
    OAA~\cite{jiang2019integral}    & \checkmark & 63.1\% & 62.8\% \\
    ICD~\cite{fan2020learning} & \checkmark & \textbf{64.0\%} & 63.9\% \\
    BES~\cite{chen2020weakly} & - & 60.1\% & 61.1\% \\
    \rowcolor{Gray} Ours (learnable) & \checkmark & 63.5\% & \textbf{64.5\%} \\
    \rowcolor{Gray} Ours (non-learnable) & \checkmark & 63.6\% & 64.4\% \\
    
    \hline
    \multicolumn{4}{c}{\textbf{Segmentation Network : DeepLab-Large-FOV (ResNet-101)}} \\
    \hline
    MCOF~\cite{wang2018weakly}      &  \checkmark & 60.3\% & 61.2\% \\
    %DCSP~\cite{chaudhry2017discovering} & \checkmark & 60.8\% & 61.8\% \\
    SeeNet~\cite{hou2018self}    & \checkmark & 63.1\% & 62.8\% \\
    AffinityNet~\cite{ahn2018learning} & -        & 61.7\% & 63.7\% \\
    FickleNet~\cite{lee2019ficklenet}  & \checkmark & 64.9\% & 65.3\% \\
    RRM~\cite{zhang2020reliability}     &  -        & 66.3\% & 65.5\% \\
    OAA~\cite{jiang2019integral}        & \checkmark & 65.2\% & 66.4\% \\
    ICD~\cite{fan2020learning} & \checkmark & \textbf{67.8\%} & \textbf{68.0\%} \\
    \rowcolor{Gray} Ours (learnable)  &  \checkmark & 66.5\% & 67.5\% \\
    \rowcolor{Gray} Ours (non-learnable) & \checkmark & 66.8\% & 67.4\% \\
    
    \hline
    \multicolumn{4}{c}{\textbf{Segmentation Network : DeepLab-ASPP (ResNet-101)}} \\
    \hline
    DSRG~\cite{huang2018weakly}     & \checkmark & 61.4\% & 63.2\% \\
    BES~\cite{chen2020weakly} & - & 65.7\% & 66.6\% \\
    SGAN~\cite{yao2020saliency}     &  \checkmark & 67.1\% & 67.2\% \\
    \rowcolor{Gray} Ours (learnable)  & \checkmark &  70.4\% & 70.7\%\\
    \rowcolor{Gray} Ours (non-learnable)  &  \checkmark & \textbf{71.2\%} & \textbf{71.4\%} \\
    
    \hline
  \end{tabular}
  \end{adjustbox}
  \caption{
    Comparison of state-of-the-art weakly-supervised semantic segmentation methods on the Pascal VOC 2012 dataset. $\mathcal{S}$ means the saliency map is used for extra guidance.
  }
  \label{tab:sota}
  \vspace{-1mm}
\end{table}
%------------------------------------------------------------------------------------

\subsection{State-of-the-arts Comparison}

We compare our approach (DRS) with other state-of-the-art weakly-supervised semantic segmentation methods that use only image-level labels as supervision.
For comparison, we apply the DRS module to all layers of the classification network and perform refinement learning.
We report the performances of both learnable controller and non-learnable controller with $\delta=0.55$.
Table \ref{tab:sota} shows the mIoU performance comparison on the PASCAL VOC 2012 validation set and test set.
We fairly compare the performance of each of the three architectures of the semantic segmentation network with other works using the same network.
Note that $\mathcal{S}$ in Table \ref{tab:sota} indicates whether the saliency map is used as extra guidance.

As shown in Table \ref{tab:sota}, DRS outperforms erasing-based methods ($e.g.,$ AE-PSL~\cite{wei2017object}, GAIN~\cite{li2018tell}, SeeNet~\cite{hou2018self}), showing that suppression is more effective than erasing.
Compared to the recent state-of-the-art methods, we achieve competitive performance despite our simplicity.
In contrast to some works ($e.g.,$ DSRG~\cite{huang2018weakly}, FickleNet~\cite{lee2019ficklenet}, AffinityNet~\cite{ahn2018learning}, BES~\cite{chen2020weakly}) where CRF in the training stage slows down the training process, our method does not apply CRF during learning, thereby achieving high performance with short training time. 
Although ICD~\cite{fan2020learning} achieves higher mIoU scores using an intra-class discriminator approach for separating foreground and background within the same image-level class, it requires a careful training strategy for stable optimization.
Unlike these methods, our approach enables fast and stable training procedure and is the simplest and the most effective way to achieve high segmentation performance.

The highlighted rows in Table \ref{tab:sota} show that the learnable and non-learnable controllers are both effective, with only a marginal difference in performance.
As mentioned in the analysis section, there is a trade-off between the two controller types, so we can choose depending on the situation.

%------------------------------------------------------------------------------------

%------------------------------------------------------------------------------------

\section{Conclusion}
In this paper, we propose a novel approach called DRS for enlarging the object regions highlighted by localization maps.
DRS propagates the initial attention to non-discriminative regions, generating dense localization maps.
The main advantage of our approach is that it is intuitive, efficient, and easily applicable to any classification network.
Together with refinement learning, our proposed method successfully generates dense segmentation labels that cover the entire target objects.
When applied to a weakly-supervised segmentation task, it achieves 71.4\% mIoU on pascal VOC segmentation benchmark using only image-level labels as weak supervision.

\bibliography{ms}
\bibliographystyle{aaai21}

\end{document}